\documentclass[11pt]{article}
\usepackage{fullpage}
\usepackage{mathptmx}
\usepackage[hidelinks=true]{hyperref}

\usepackage[numbers,sort&compress]{natbib}
\usepackage{xspace}
\usepackage[normalem]{ulem}
\usepackage{amsmath,amssymb,url,subfig}
\usepackage{algorithm}
\usepackage[noend]{algpseudocode}
\usepackage{booktabs}
\usepackage{tabularx}

\newcommand{\lmkl}{\texttt{LMKL}\xspace}
\newcommand{\clmkl}{\texttt{C-LMKL}\xspace}
\newcommand{\smkl}{\texttt{SwMKL}\xspace}
\newcommand{\samkl}{\texttt{SAMKL}\xspace}
\newcommand{\ldmkl}{\texttt{LD-MKL}\xspace}
\newcommand{\gmkl}{\texttt{SPG-GMKL}\xspace}
\newcommand{\reals}{\mathbb{R}}
\newcommand{\innerprod}[2]{\left\langle #1, #2 \right\rangle}

\usepackage{mathtools}
\mathtoolsset{showonlyrefs}

\DeclareMathOperator{\softmax}{softmax}

\renewcommand{\vec}[1]{\textbf{#1}}

\begin{document}

\title{\Large A Unified View of Localized Kernel Learning\thanks{This research
    was funded in part by the NSF under grants IIS-1251049 and CNS-1302688.}}
\author{John Moeller\thanks{School of Computing, University of
    Utah. \textbf{Email:} \textsl{\{moeller, sarath, suresh\}@cs.utah.edu}} \\
\and
Sarathkrishna Swaminathan\footnotemark[2]\\
\and
Suresh Venkatasubramanian\footnotemark[2]}

\date{}

\maketitle

\begin{abstract} 
  Multiple Kernel Learning, or MKL, extends (kernelized) SVM by
  attempting to learn not only a classifier/regressor but also the best kernel
  for the training task, usually from a combination of existing kernel
  functions.  Most MKL methods seek the combined kernel that performs best over
  \emph{every} training example, sacrificing performance in some areas to seek a
  global optimum.  \emph{Localized} kernel learning (LKL) overcomes this
  limitation by allowing the training algorithm to match a component kernel to
  the examples that can exploit it best.  Several approaches to the localized
  kernel learning problem have been explored in the last several years.  We
  unify many of these approaches under one simple system and design a new 
  algorithm with improved performance. We also develop enhanced versions of
  existing algorithms, with an eye on scalability and performance.
\end{abstract}

\section{Introduction}
\label{sec:intro}

Kernel-based learning algorithms require the user to specify a kernel that
defines the \emph{shape} of the underlying data space. \emph{Kernel learning} is
the problem of learning a kernel from the data, rather than providing one by
fiat. Most approaches to kernel learning assume some structure to the kernel
being learned: either as an explicit linear Mahalanobis representation, or as
some finite combination of set of fixed kernels. This latter class of approaches
is often called \emph{multiple kernel learning}. 

\begin{figure*}[h]
  \centering
  \subfloat[Classifier produced by global MKL with 118 support points.\label{fig:global}]{\includegraphics[width=0.5\columnwidth]{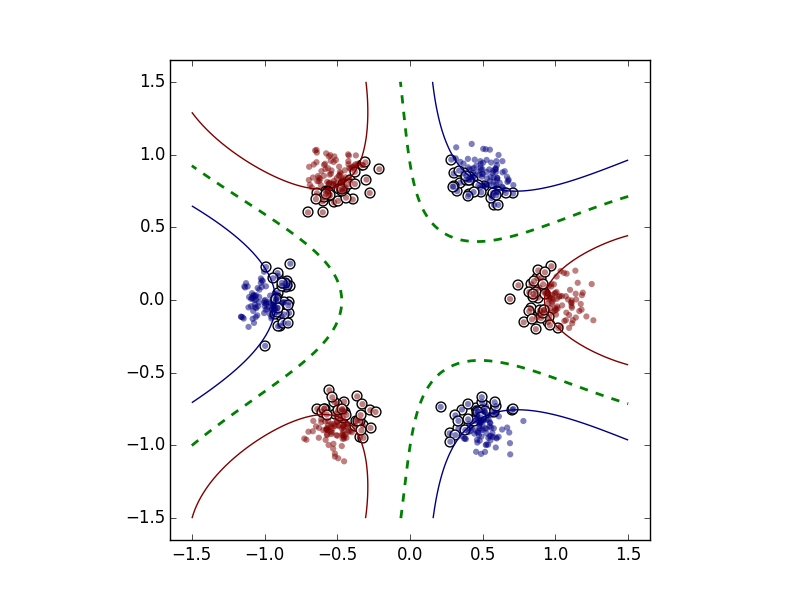}}
  \subfloat[Classifier produced by \ldmkl with 20 support points.\label{fig:local}]{\includegraphics[width=0.5\columnwidth]{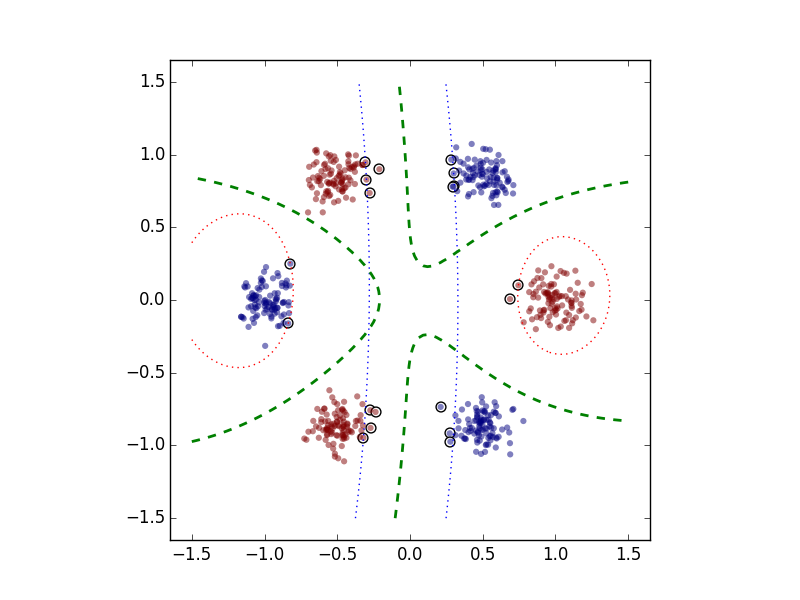}}
  \caption{Illustration of the difference between local and global multiple
    kernel learning. In each example, the classifier is built from two kernels,
    one quadratic and one Gaussian. Points from the two classes are colored blue and
    red (with transparency as a hint towards density). The decision boundary is
    marked in green and the margin boundaries are in the appropriate
    colors for the global case. For the local case, the margins of each kernel are plotted with dotted lines, red for Gaussian and blue for quadratic. Support points are indicated by black circles around points. Note
    that the classifier uses a soft-margin loss and so support points may not be
  exactly on the margin boundary. The global version has 118 support points, while the local version has only 20.}
  \label{fig:example}
\end{figure*}

While multiple kernel learning has been studied extensively and has had success
in identifying the right kernel for a given task, it is expressively limited
because each kernel has influence over the entire data space. Consider an
example of a binary classification task, depicted in
Figure~\ref{fig:example}. On the left side we show the results of classifying
the data with a global MKL method (here, the \textsc{Uniform}  method of
\citet{DBLP:conf/nips/CortesMR09}) and on the right side we show the results of
classification with our new proposed  method \ldmkl. Because the global method requires that each kernel be used to
classify each point in the same way, the decision boundary is not as flexible
and many more support points are required. 

Motivated by this, a few directions have been proposed to build \emph{localized} kernel
learning solutions. \citet{DBLP:journals/pr/GonenA13} introduced the idea
of a learned \emph{gating} function that modulated the influence of a kernel on
a point (\lmkl). 
\citet{DBLP:journals/corr/LeiBDK15} observed that \lmkl uses a non-convex optimization and suggested using a
probabilistic clustering to generate part of the gating function beforehand, in order to obtain a convex
optimization and thus prevent over-fitting and yield generalization bounds (\clmkl).
\citet{smkl} suggested a different approach to find a
gating function by looking at individual features of the input, and uses successes of the individual kernels to learn the gating function through support vector regression (\smkl). 

All of the above approaches invoke a fixed-kernel SVM subroutine 
as part of the algorithm. This is inefficient, and
prevents these methods from scaling. \clmkl does argue for a convex
formulation of the problem, but does not directly address
the problem of scaling. 

\subsection{Our Contributions.}
\label{sec:our-contributions}

We present a \emph{unified} interpretation of localized  kernel learning
that generalizes all of the approaches described above, as well as the general
multiple kernel learning formulation. This interpretation yields a new algorithm
for LKL that is superior to all existing methods. In addition, we make use of prior
work on scalable multiple kernel learning~\cite{DBLP:conf/aistats/MoellerRVS14}
as a subroutine to make existing methods for LKL scale well, improving their
performance significantly in some cases. 

Our interpretation relies on a geometric interpretation of gating functions in
terms of \emph{local} reproducing kernel Hilbert spaces acting on the data. This
interpretation also helps explain the observation above (only empirically observed
thus far) that local kernel learning methods appear to produce good classifiers
with fewer support points than global methods.

\section{Background}

We use the notation $[a\ldots b]$ to indicate a sequence of integers $i$ such that $a\leq i\leq b$.
We use bold Roman letters to indicate vectors ($\vec{x}$) and matrices ($\vec{A}$). 
Matrices are capitalized.
Because we discuss several approaches to localized MKL, and each uses a different set of notations, we choose our own convention:
\begin{itemize}
\item $i$ indexes kernel functions/spaces and the number of individual kernel spaces is $m$.
\item $j$ and $k$ index examples and the number of training points is $n$.
\item $t$ is used to indicate iterations in an algorithm.
\item The Greek letter $\kappa$ is used to indicate a kernel function. 
$\kappa_i(\vec{x}_j,\vec{x}_k)$ is the $i$th kernel function applied to training examples $\vec{x}_j$ and $\vec{x}_k$.
\end{itemize}

The $\softmax$ operator is a map $\mathbb{R}^d \to (0,1)^d$ that normalizes the input vector to the range $(0,1)$: 
\begin{equation}
  \softmax(\vec{x}) = \left( \frac{\exp{x_i}}{\sum_{k=1}^d \exp{x_k}} \right)_{i=1}^d
\end{equation}

\subsection{Prior work on localized kernel learning}
\label{sec:prior-local-kern}

Since our work unifies a number of different approaches to performing localized
multiple kernel learning, we start with a self-contained review of these
methods. 

The core idea (somewhat simplified) of kernel learning for classification is to fix a space
$\mathcal{K}$ of kernel functions $\kappa(\cdot, \cdot)$ and learn a kernel
$\kappa \in \mathcal{K}$ that best classifies the training data. The term
\emph{multiple kernel learning} comes from the fact that the space $\mathcal{K}$
is often expressed as the set of all positive combinations of a fixed set of
kernels, and thus the search for a specific kernel turns into a search for a set
of parameters $\eta_i$, one per kernel, so that the resulting discriminant
function can be written as 
\[ f(x) = \sum_{i=1}^m \eta_i \langle \mathbf{w}_i, \phi_i(\vec{x})\rangle + b \]

The rationale for \emph{localized}  kernel learning (as illustrated in
Section~\ref{sec:intro}) is to allow the weight assigned to different kernels to
vary in different parts of the data space to incorporate any local structure in
the data. 

\paragraph{Localized Multiple Kernel Learning (\lmkl).}
\citet{DBLP:conf/icml/GonenA08} were the first to propose an algorithm to solve
this problem. They called their method \emph{localized multiple kernel learning}
(\lmkl). The idea was generalize the $\eta_i$ to be functions of the data
$\vec{x}$ as well as a set of \emph{gating parameters} $\mathbf{V} \in \reals^{d \times
  m}$. 

They defined a gating function as:
\begin{equation}
  \eta (\vec{x} | \vec{V}) = \softmax(\vec{x}^\top \vec{V} + \vec{v}_{0}),
\end{equation}
where $\vec{v}_0$ is an $m$-dimensional vector of offsets\footnote{
In later works they proposed other gating functions that employed sigmoids and
Gaussian functions~\citep{DBLP:journals/pr/GonenA13}.}.

Given such a gating function, they then defined a generalized discriminant function:
\begin{equation}
  f(\vec{x}) = \sum_{i=1}^{m} \eta_i (\vec{x} | \vec{V}) \langle w_i, \phi_i (\vec{x}) \rangle + b ,
\end{equation}

Expressing the classifier function leads to a non-convex optimization involving
the parameters $V$. They then proposed solving this problem using a two-step alternating optimization algorithm, summarized in Algorithm~\ref{alg:lmkl}.

\begin{algorithm}
  \caption{\lmkl}
  \begin{algorithmic}[1]
    \Repeat
    \State Calculate $\vec{K}_\eta$, the Gram matrix of the combined kernel, with the gating functions $\eta_i$:
    \State $(K_\eta)_{jk} \gets \kappa_{\eta} (\vec{x}_j, \vec{x}_k) = \sum_{i=1}^{m} \eta_i (\vec{x}_j) \kappa_i (\vec{x}_j, \vec{x}_k) \eta_i(\vec{x}_k)$
    \State Solve canonical SVM with $\vec{K}_\eta$
    \State Update gating parameters $\mathbf{V}$ using gradient descent
    \Until{convergence}
  \end{algorithmic}
  \label{alg:lmkl}
\end{algorithm}

The complexity of the overall algorithm is dominated by the time to perform the
canonical SVM. Other variants of this basic framework include \citet{yang2009group},
which allows gating functions to operate on groups of points, and \citet{han2012probability}
which incorporates a gating function based on pair-wise similarities inferred
from a kernel density estimate for each kernel.

\paragraph{Convex LMKL (\clmkl).}

More recently, \citet{DBLP:journals/corr/LeiBDK15} noted the non-convex nature
of the above objective function. In order to avoid the tendency of such
functions to overfit to the training data, they proposed an alternate
\emph{convex} formulation of the localized multiple kernel learning problem. The
central idea of their approach is to first construct a soft clustering of the
data, represented by a soft assignment function $c_\ell(x_j)$ that associates point
$x_j$ with cluster $\ell$. Next, they define parameters $\beta_{\ell i}$ that associate each of
$m$ kernels with each cluster $\ell$: in effect, the soft clustering fixes the
locality they wish to exploit, and the $\beta_{\ell i}$ then allow them to use
different kernel combinations. 

The resulting optimization is convex, assuming that the loss function is
convex. This allows them to obtain generalization bounds as well as good
prediction accuracy in practice. The optimization itself proceeds as a two-stage
optimization: the first stage invokes a standard SVM solver to find the best
weight vectors given the $\beta_{\ell i}$ and the second stage optimizes
$\beta_{\ell i}$ for given weights. This latter stage can in fact be solved in
closed form. Thus, as with \lmkl, the term dominating the computation time is
the use of an SVM solver.

\paragraph{Success-Based Locally-Weighted Kernel Combination (\smkl)}

\citet{smkl} introduced \smkl as a way to localize kernel learning in a different manner. 
Their method is to analyze each kernel for its success on the input data, then construct a gating function based on smoothing the success with a regression, summarized in Algorithm~\ref{alg:smkl}.

\begin{algorithm}
  \caption{\smkl}
  \begin{algorithmic}[1]
    \ForAll {$i\in [1..m]$}
    \State Train classifier $f_i : \mathbb{R}^d \to \{-1,1\}$ with kernel $\kappa_i$
    \State Train regressor $g_i : \mathbb{R}^d \to (0,1)$ with $(\vec{X},\delta(\vec{y}, f_i(\vec{X})))$
    \EndFor
    \State Train classifier using $$\kappa(\vec{x}_j, \vec{x}_k) = \frac {\sum_{i=1}^{m} g_i (\vec{x}_j) \kappa_i (\vec{x}_j, \vec{x}_k) g_i(\vec{x}_k)} {\sum_{i = 1}^{m} g_i (\vec{x}_j) g_i (\vec{x}_k)}$$
  \end{algorithmic}
  \label{alg:smkl}
\end{algorithm}

Its complexity is controlled by the initial SVM computations, the different support
vector regression operations, as well as the final SVM calculation on the
combined kernel function. 
The experimental approach in~\cite{smkl} is to separate each kernel by feature -- 
essentially creating individual kernels for each combination of kernel and feature and then combining them. 
When testing with this algorithm, we had much better success when using a kernel on all features.

\paragraph{Sample-Adaptive Multiple Kernel Learning (\samkl).}

An alternate approach employed by \citet{samkl} is to separate out the
assignment of kernels to points and the weights associated with the kernels. In
their formulation, which they describe as \emph{sample-adaptive multiple kernel
  learning}, they introduce latent \emph{binary} variables to decide whether a
particular kernel should operate on a particular point or not. Each point is
therefore mapped to a single point in the product of the feature spaces defined
by the given kernels. Now they run a two-stage alternating optimization: in the
first stage, given fixed values of the latent variables, they solve a multiple
kernel learning problem for the different subspaces simultaneously, and then
they run an integer program solver to obtain new values of the latent
variables. Note that each step of the iteration here involves costly operations
(an MKL solver and an integer program solver) in comparison with the SVM solvers
in the other approaches.

\section{A unified view of localized kernel learning}
\label{sec:clmkl}

One of the contributions of this work is a unified perspective that integrates
these different approaches and also helps explain the somewhat paradoxical fact
that localized multiple kernel often yields classifiers with \emph{fewer}
support points than standard multiple kernel learning methods. 

\subsection{Localization via Hilbert subspaces}\label{sec:lmkl-as-hilbert}

Consider the following generalized and gated kernel $\kappa_\gamma$ defined as:
\begin{equation}
  \kappa_\gamma(\vec{x}, \vec{x}') =  \sum_{i=1}^{m} \gamma_i(\vec{x},\vec{x}') \kappa_i (\vec{x}, \vec{x}'),
\end{equation}
where $\gamma_i : \reals^d\times\reals^d \to [0,1]$ is a ``gating function.'' 

We call $\gamma_i$ \emph{separable} if it decomposes into a product of a function with itself, i.e. if $\gamma_i(\vec{x},\vec{x}') = \eta_i(\vec{x})\eta_i(\vec{x}')$, where $\eta_i : \reals^d \to [0,1]$.
For the rest of this section, we only consider separable gating functions.
We also make two additional assumptions for all $\vec{x}\in\reals^d$: 
(1) $\sum_{i=1}^m \eta_i(\vec{x}) = 1$, and (2) $\eta_i(\vec{x}) \geq 0 \ \forall i\in[1..m]$.

\paragraph{The RKHS of a localized kernel.} Consider the Gram matrix $\vec{H}_i$
of $\gamma_i$: specifically the $n\times n$ matrix $\vec{H}_i$ whose
$(j,k)^{\text{th}}$ entry is $\gamma_i(\vec{x}_j, \vec{x}_k)$ (we will  refer to this later as the \emph{gating matrix}). 
If $\gamma_i$ is separable, then we know that $\vec{H}_i$ is positive definite, because it can be expressed as the outer product of a vector with itself ($\vec{H}_i = \boldsymbol{\eta}^\top\boldsymbol{\eta}$).
Defining $\vec{K}_i$ as the Gram matrix of the kernel $\kappa_i$, it is now easy
to see that we can write the Gram matrix of the kernel  $\kappa_\gamma$ as the
matrix $\sum_i \vec{H}_i\circ\vec{K}_i$ (where $\circ$ denotes the Schur product).

In the separable case, since both $\vec{H}_i$ and $\vec{K}_i$ are positive definite, so is $\vec{H}_i\circ\vec{K}_i$ by the Schur product theorem.
Therefore $\gamma_i(\vec{x},\vec{x}') \kappa_i (\vec{x}, \vec{x}')$ is a kernel function, and the corresponding lifting map is $\eta_i(\vec{x})\Phi_i(\vec{x})$.

We know that a positive linear combination of kernel functions is itself a
kernel function and induces a product reproducing kernel Hilbert space (RKHS) that is  a
simple Cartesian product of all the individual Hilbert spaces. The inner product
of this space is just the sum of all the individual inner products. Thus the
kernel $\kappa_\gamma$ has a natural feature space as the product of the
individual feature spaces. 

\paragraph{Localization.} 
This framework now allows us to provide a geometric intuition for why localized
kernel learning might be able to reduce the number of required support points. 
Suppose that $\eta_i(\vec{x}) = 0$. 
This implies that $\innerprod{\eta_i(\vec{x})\Phi_i(\vec{x})}{\eta_i(\vec{x}')\Phi_i(\vec{x}')}$ is always 0. 
Because the $i^{\text{th}}$ RKHS is one component of the product RKHS, this
means that $\eta_i(\vec{x})\Phi_i(\vec{x})$ lies in some subspace perpendicular
to this RKHS. 

Furthermore, suppose that $\eta_i(\vec{x}) = 1$. 
By our assumptions that $\sum_{i=1}^m \eta_i(\vec{x}) = 1$ and that $\eta_i$ is non-negative, this means that $\eta_i(\vec{x})\Phi_i(\vec{x})$ is absent from \emph{every other} RKHS in the product. 
Therefore $\eta_i(\vec{x})\Phi_i(\vec{x})$ lies exclusively in the $i$-th RKHS.

This partitioning behavior is advantageous, because it is much simpler to find decision boundaries within the individual RKHS components rather than trying to find one that will work for all at the same time.
The decision hyperplane in the product RKHS will be the unique hyperplane that intersects all the subspaces in their respective decision boundaries.

Depending on the gating function, there will of course be some training examples
that are ``confused'' about what subspace to lie in.  Therefore we wish to pick
a set of gating functions that reduces this confusion.  The \emph{crucial}
property of the gating function $\gamma_i$ and the gating matrix $\vec{H}_i$ is
that they are separable.  With the separability constraint, we need only find a
set of one-dimensional functions that works for the training data\footnote{ If
  the gating function is not separable, but is decomposable into a positive
  linear combination of a fixed-size set of separable functions, then the
  partitioning is still possible -- see Section~\ref{sec:clmkl} below, under
  ``\clmkl''.  }.

\subsection{Gating and optimization}
\label{sec:gating-optimization}

The localized MKL algorithms described above (and in fact virtually all
localized kernel learning algorithms) can be placed in the framework we have
just described, thus explaining in a broader context how their localization
works. The specifics differ on how the function $\kappa_\gamma$ is
generated: 
\begin{enumerate}
\item \textbf{Gating}: 
  Each algorithm has a gating function $\gamma_i(\vec{x},\vec{x}')$ for every kernel function $\kappa_i$. 
  Recall that the gating function simply controls the degree to which a kernel responds to a particular point.
\item \textbf{Optimization}:
  Each algorithm also has an optimization behavior, that either generates or tunes each $\gamma_i$.
\end{enumerate}

\subsubsection*{\lmkl:} 
\begin{itemize}
\item \textbf{Gating}: The gating function is separable, and $\eta(\vec{x}) =
  (\eta_1(\vec{x}), \ldots, \eta_i(\vec{x}), \ldots) = \softmax(\vec{x}^\top \vec{V} + \vec{v}_{0})$.
\item \textbf{Optimization}: Alternating optimization using an SVM solver to
  find the kernel support points and stochastic gradient descent to find the
  parameters $\vec{V}$, $\vec{v}_0$. 
\end{itemize}
    
\subsubsection*{\clmkl:}
\begin{itemize}
\item \textbf{Gating}: The gating function is separable, but not directly. 
It is equal to $\sum_{r = 1}^{\ell} \beta_{ir} c_r (\vec{x}) c_r(\vec{x}')$,
where $\beta_{ir}\geq 0$ is the weight with which kernel $i$ influences points
associated with  cluster $r$, and $c_r$ is the (pre-computed) likelihood of
$\vec{x}$ falling into cluster $r$.  
  
  Since $\gamma_i$ decomposes into a linear combination $\beta_{ir} c_r (\vec{x}) c_r(\vec{x}')$, we can apply Section~\ref{sec:lmkl-as-hilbert} to \clmkl.
  In \clmkl we replicate each kernel $\ell$ times (once for each $c_r$) and give each its own weight $\sqrt{\beta_{ir}}$.
\item \textbf{Optimization}: The parameters $\beta_{ir}$ are learned through
  (convex) optimization and the functions $c_r$ are generated through $\ell$ different clusterings. 
\end{itemize}

\subsubsection*{\smkl:} 
\begin{itemize}
\item \textbf{Gating}: The gating function is not separable in this case, because the $\gamma_i$ are normalized \emph{pairwise}. 
  $\gamma_i(\vec{x}, \vec{x}') = {g_i (\vec{x}) g_i(\vec{x}')}/{Z(\vec{x},\vec{x}')}$, where $Z(\vec{x},\vec{x}') = {\sum_{i=1}^{m} g_i(\vec{x}) g_i(\vec{x}')}$, and $g_i$ are the SVR-generated functions.

  Note While $\kappa_\gamma$ may be positive definite, its individual terms are
  very unlikely to be so. It is therefore not clear whether this algorithm in
  its unmodified form can be placed in our unified context. We explore this
  issue in greater depth in the next section. 
\item \textbf{Optimization}: The gating functions $g_i$ are generated using SVR
  from $\vec{X} \times \delta(\vec{y}, \hat{\vec{y}}_{i})$.
\end{itemize}

\subsubsection*{\samkl}
\label{sec:sample-adapt-mult}

\begin{itemize}
\item \textbf{Gating:} $\eta_i(\vec{x})$ is a \emph{binary-valued} function that
  decides if kernel $i$ should be used for point $\vec{x}$. 
\item \textbf{Optimization:} The optimization is an alternating optimization
  between the gating function and the kernel parameters. Because the $\eta_i$ are binary-valued, a further
  \emph{multiple} kernel learning step is required to determine kernel weights
  and support vectors for the classifier, and the gating parameters are learned
  with an \emph{integer programming} solver. 
\end{itemize}

\subsubsection*{Global (``classic'') MKL: }
\begin{itemize}
\item \textbf{Gating}: $\eta_i (\vec{x}) =  \sqrt{\mu_i}$, where $\mu_i\geq 0$ is constant for every kernel, that is, does not change relative to each point.
\item \textbf{Optimization}: The $\mu_i$ can be optimized using several methods
  including stochastic gradient descent, multiplicative weight updates and
  alternation. 
\end{itemize}

\section{\ldmkl: A new algorithm for localized kernel learning}

Viewing the algorithms for localized kernel learning in a common framework illustrates both their commonalities and their weaknesses. With the exception of \smkl, all the approaches make use of a two- (or three-) stage optimization of which \texttt{LibSVM} is one component. As we shall see in our experiments, this renders these methods quite slow and not easy to scale. \smkl on the other hand avoids this problem by doing single SVM calculations for each kernel and then combining them into a single larger kernel. This improves its running time, but makes it incur a large memory footprint in order to build a classifier for the final kernel. 

We now present a new approach, inspired by \smkl, that addresses these concerns. Our method, which we call \ldmkl (\emph{localized decision-based multiple kernel learning}), fits into the unified framework for localized kernel learning via the use of local Hilbert spaces, avoids the large memory footprint of \smkl, and also scales far more efficiently than the other multi-stage optimizations.

We start by observing that the first steps of Algorithm~\ref{alg:smkl} give us a classifier $f_i$ and a gating function $g_i$.
The function $f_i$, since it is an SVM decision function, can be formulated as
\begin{equation}
  f_i(\vec{x}) = \sum_{j=1}^n \alpha_{ij} y_j \kappa_i(\vec{x}_j,\vec{x}).
\end{equation} 
Note that $\alpha$ has an additional index to indicate which kernel we trained the classifier against.
Suppose we modify this function to incorporate the gating function $g_i$\footnote{As discussed in the previous section, we assume that the gating functions have been normalized so that (1) $\sum_{i=1}^m g_i(\vec{x}) = 1$ and (2) $g_i(\vec{x}) \geq 0 \ \forall i\in[1..m]$.
}:
\begin{align}
  \overline{f}_i(\vec{x}) &= \sum_{j=1}^n \alpha_{ij} y_j g_i(\vec{x}_j) \kappa_i(\vec{x}_j,\vec{x}).\label{eq:1}
\end{align}

$\overline{f}_i$ is the SVM prediction function, but where each support point $\alpha_{ij}$ is weighted by its gating value.
We can now construct a weighted vote using these functions.
We combine the output of each $\overline{f}_i$, apply $\tanh$\footnote{%
We use $\tanh(\overline{f}_i(\vec{x}))$ instead of the sign of $\overline{f}_i(\vec{x})$ so that uncertain classifications (i.e., kernels with resulting values of $\overline{f}_i(\vec{x})$ near $0$) don't pollute the vote with noise.
}, and weight by $g_i$:
\begin{align}
  f(\vec{x}) &= \sum_{i=1}^{m} g_i (\vec{x}) \tanh(\overline{f}_i(\vec{x}) )\label{eq:2}
\end{align}

Algorithm~\ref{alg:ldmkl} contains the listing of this procedure. 
Note that we \emph{retrain} each classifier on the subset of the data where the corresponding gating function is significant (i.e. is greater than $1/m$).
This reduces the support points considerably because the classifier is retrained only on points that it classified well.

\begin{algorithm}
  \caption{\ldmkl}
  \begin{algorithmic}[1]
    \ForAll {$i\in [1..m]$}
    \State Train classifier $f_i : \mathbb{R}^d \to \{-1,1\}$ with kernel $\kappa_i$
    \State Train regressor $g_i : \mathbb{R}^d \to (0,1)$ with $(\vec{X},\delta(\vec{y}, f_i(\vec{X})))$
    \EndFor
    \State Normalize regressors $g_i$ with $\softmax$
    \ForAll {$i\in [1..m]$}
    \State Retrain classifier $f_i$ on $(\vec{X},\vec{y})_{g_i(\vec{x})>1/m}$
    \EndFor
    \State Compute each decision function using \eqref{eq:1}
    \State Classify inputs using sign of \eqref{eq:2}
  \end{algorithmic}
  \label{alg:ldmkl}
\end{algorithm}

If commonly-used kernels are employed (such as linear, polynomial, or Gaussian kernels), then this method can take advantage of optimizations that exist in, e.g., LibSVM to train the classifiers and regressors quickly. The training step is over after the regressors are computed and normalized.

It is easy to see that \ldmkl has the desired \emph{gating} behavior with separable gating functions. The optimization step is as before, but without needing to consult a final SVM solver.

\section{Experiments}

Our experiments will seek to validate two main claims: first that \ldmkl is indeed
superior to prior localized kernel learning methods, and secondly that there is
demonstrable reduction in the number of support points when using localized
methods. 

\paragraph{Scalability.} In addition, we will also investigate ways to make
existing localized methods more scalable. As noted, with the exception of \smkl,
all approaches use a multi-stage iterative optimizer of which one step is
an SVM solver. We instead make use of a multiplicative-weight-update-based
solver developed by \citet{DBLP:conf/aistats/MoellerRVS14}. This method has a
much smaller memory footprint and uses a lightweight iteration that also yields
sparse support vectors. While this solver was designed for \emph{multiple}
kernel learning, it is easily adapted as an SVM solver.

\paragraph{Data Sets.} Table~\ref{tab:t0-5} contains information about the various datasets that we
test with. All of these sets are taken from the \texttt{libsvm} repository at
\url{https://www.csie.ntu.edu.tw/~cjlin/libsvmtools/datasets/binary.html}. 

\begin{table}[htbp]
  \centering
  \begin{tabular}{lrr}
\toprule
    Dataset       & Examples & Features \\ \midrule
    Breast Cancer & $683   $ & $10$     \\
    Diabetes      & $768   $ & $8$      \\ 
    German-Numeric  & $1000  $ & $24$     \\ 
    Liver         & $345   $ & $6$      \\ 
    Mushroom      & $8192  $ & $112$    \\
    Gisette       & $6000  $ & $5000$   \\
    Adult         & $32561$ & $123$    \\ \bottomrule
  \end{tabular}
  \caption{Datasets for comparison of \lmkl, \smkl, and \clmkl}
  \label{tab:t0-5}
\end{table}

\paragraph{Methodology.} In each of the experiments we partition the data randomly between $75\%$ train and $25\%$ test examples. 
Unless otherwise indicated, we repeat each partition 100 times and average the run time and the accuracy.
In all experiments where we measure accuracy, we use the proportion of correctly classified points.
Where possible, we also report the standard deviation of all measured values in parentheses.
Superior values are presented in bold when the value minus the standard deviation is greater than all the other values plus their respective standard deviations.

In each experiment where we used a standard SVM solver, we used LibSVM~\cite{CC01a} via \texttt{scikit-learn}~\cite{scikit-learn}.
We use the default LibSVM parameters (e.g., tolerance), and vary them only for changing specific kernels and passing specific kernel parameters.
We use $C=1.0$ and for Gaussian kernels, a range of $\gamma$ from $2^{-4}$ to $2^4$ are tried and the best accuracy observed is used.

\paragraph{Implementations.}

For \lmkl, we took \textsc{Matlab} code provided by
Gonen\footnote{\url{http://users.ics.aalto.fi/gonen/icml08.php}} and converted
it to \texttt{python} to have a common platform for comparison. This code
included an SMO-based SVM solver which we converted as well. We verified
correctness of intermediate and final results between the two platforms before
running our experiments. For \smkl and \ldmkl, we used the SVM and SVR solvers from
\texttt{scikit-learn}. For \clmkl, as prescribed by
\citet{DBLP:journals/corr/LeiBDK15}, we used a kernel $k$-means preprocessing
step with a uniform kernel and three clusters. For large data sets, kernel
$k$-means is very slow, and so we used a streaming method proposed by
\citet{chitta2011approximate} that runs the clustering algorithm on a sample (of
size $1000$ in our experiments) and then estimates probabilities for the remaining points.
The \emph{global} kernel learning methods we used were \textsc{Uniform}, which
merely averages all kernels, \gmkl
\cite{DBLP:conf/kdd/JainVV12}\footnote{\url{http://www.cs.cornell.edu/~ashesh/pubs/code/SPG-GMKL/download.html}}
and \textbf{MWUMKL} \cite{DBLP:conf/aistats/MoellerRVS14}. 

\begin{table*}[t]
\small
  \centering
  \begin{tabular}{lcccc}
\toprule
             & \lmkl              & \smkl            & \ldmkl           & \clmkl         \\ \midrule
    Breast   & 96.58 \% (1.35 \%) & 97.1\% (1.1\%)   & 97.1\% (1.2\%)   & 96.7\% (1.1\%) \\
    Cancer   & 122 s (8.9 s)      & 0.15 s (2.1 ms)  & \textbf{0.14 s} (2.36 ms) & 28.7 s (80 ms) \\ \hline
    Diabetes & 74.71\% (3.07\%)   & 77.0\% (2.7\%)   & 76.7\% (2.56\%)  & 76.4\% (2.4\%) \\ 
             & 157.6 s (34 s)     & \textbf{0.18 s} (1.4 ms)  & 0.24 s (3.58 ms) & 36.8 s (32 ms)  \\ \hline
    German-  & 70.78\% (2.85\%)   & 75.7\% (2.4\%)   & 75.84\% (2.51\%) & 76.8\% (1.6\%) \\ 
    Numeric  & 216 s (22 s)       & \textbf{0.27 s} (3.8 ms)  & 0.38 s (3.56 ms) & 69.6 s (20 ms)  \\ \hline
    Liver    & 62.49\% (5.92\%)   & 69.3\% (5.0\%)   & 65.19\% (5.81\%) & 57.7\% (5.1\%) \\ 
             & 35.4 s (7.2 s)     & \textbf{0.1 s} (1.4 ms)   & 0.83 s (1.3 ms)  & 7.4 s (129 ms) \\ \hline
    Mushroom & 99.99\% (0.0\%)    & 99.9\% (0\%)     & 100.0\% (0.0\%)  & 100\% (0.0\%)  \\
             & 17.27 m (1.2 m)    & 14.57 s (1.2 s)  & \textbf{3.1 s}  (0.3 s) & 2.43 h (12.6 m)\\ \hline
    Gisette  & 97.22\% (0.34\%)   & 97.06\% (0.35\%) & 96.88\% (0.46\%)  & 96.5\% (0.28\%) \\ 
             & 48.6 m (2.8 m)     & 4.54 m (0.72 m)  & 4.0 m (0.06 m) & 3.4 h (9.24 m)  \\ \hline 
    Adult    &  -                 & 84.6\% (0.37\%)  & 84.78\% (0.4\%) & 84.65\% (0.83\%) \\ 
    Income   & -                  & 6.65 m (1.2 m)   & 6.52 m (0.14 m) &  7.5 h (14.3 m) \\ \bottomrule
  \end{tabular}
  \caption{Accuracies and running times for various datasets and methods, using \textbf{LibSVM} as the SVM solver. 
    Numbers in parentheses are standard deviations. 
    For the first four data sets, numbers are averaged over $100$ runs. 
    For the last three larger data sets, numbers are averaged over $20$ runs. 
    Values which are significantly superior to that of other methods are typeset in bold.}
  \label{tab:t0}
\end{table*}

\paragraph{Hardware.}
All experiments were conducted on Intel\textregistered\ Xeon\textregistered\ E5-2650 v2 CPUs, 2.60GHz with 64GB RAM and 8 cores.

\subsection{Evaluating \ldmkl.}
\label{sec:evaluating-ldmkl}

We start with an evaluation of \ldmkl in Table~\ref{tab:t0}. 
In each row, we present accuracy and timing (numbers in parentheses are standard deviations). 
As we can see, for small datasets, \smkl is the fastest method, but for larger datasets \ldmkl is the fastest.
In comparison with \lmkl and \clmkl, \smkl and \ldmkl are considerably faster. 
This speedup is obtained without any significant loss in accuracy: 
in all cases, the accuracy of \ldmkl is either the best or is less than optimal in a statistically insignificant way.

\begin{table*}[htbp]
\small
  \centering
  \begin{tabular}{lcccc}
\toprule
             & \lmkl              & \smkl             & \ldmkl                  & \clmkl          \\ \midrule
    Breast   & 97.08 \% (1.1 \%)  & 96.42\% (1.6\%)   & 93.4\% (2.2\%)          & 90.4\% (2.4\%)  \\
    Cancer   & 0.18 s (4.3 ms)    & 0.18 s (1.2 ms)   & 0.63 s (9.3 ms)         & 5.6 s  (122 ms) \\ \hline
    Diabetes & 73.19\% (3.39\%)   & 76.63\% (2.9\%)   & 77.0\% (3.4\%)          & 71.1\% (10\%)   \\ 
             & \textbf{0.27 s} (18 ms)     & 0.29 s (3.8 ms)   & 0.48 s (46 ms) & 7.2 s (32 ms)   \\ \hline
    German-  & 70.07\% (3.1\%)    & 72.2\% (3.29\%)   & 73.0\% (3.8\%)          & 73.4\% (4.1\%)  \\ 
    Numeric  & 0.63 s (43 ms)     & 0.62 s (10.2 ms)  & 1.0 s (55 ms)             & 16.3 s (101 ms) \\ \hline
    Liver    & 56.82\% (6.53\%)   & 59.63\% (10.47\%) & 58.8\%  (8.0\%)         & 49.7\% (6.3\%)  \\ 
             & 0.13 s (5 ms)      & \textbf{0.11 s} (3.4 ms)   & 0.3 s (6.5 ms) & 1.45 s (106 ms) \\ \hline
    Mushroom & 99.87\% (0.1\%)    & 99.9\% (0\%)      & 99.9\% (0.1\%)          & 98.8\% (0.24\%) \\
             & 24.4 s (0.36 s)    & \textbf{21.3 s} (0.2 s)    & 53.0 s (0.2 s)   & 31.4 m (1.2 m)  \\ \hline
    Gisette  & \textbf{97.28\%} (0.4\%)    & 69.96\% (2.01\%)  & 92.2\% (0.8\%) & 90.26\% (1.2\%) \\ 
             & \textbf{8.2 m} (0.18 m)     & 8.91 m (0.44 m)   & 29.0 m (10 s)  & 28.5 m (53.1 s) \\ \hline 
    Adult    & 57.4\% (5.31\%)    & 83.96\% (0.61\%)  & 80.2\% (0.8\%)          & 84.65\% (0.35\%) \\ 
    Income   & 9.4 m (0.6 m)      & 9.1 m (6.8 s)    & 12.3 m (14.3 s)          & 47.65 m (2.46 m) \\ \hline
  \end{tabular}
  \caption{Accuracies and running times for various datasets and methods, using \textbf{MWUMKL} as the SVM solver.  
    Numbers in parentheses are standard deviations. 
    For the first four data sets, numbers are averaged over $100$ runs. 
    For the last three larger data sets, numbers are averaged over $20$ runs. 
    Values which are significantly superior to that of other methods are typeset in bold.}
  \label{tab:t0-1}
\end{table*}

\subsection{Scaling}

As we can see in Table~\ref{tab:t0}, \lmkl and \clmkl run very slowly as the
data complexity increases (dimensions or number of points), and the primary
bottleneck is the repeated invocation of an SVM solver. As described above, we
replaced the SVM solver with a single-kernel version of \textbf{MWUMKL} and
studied the resulting performance. 

Table~\ref{tab:t0-1} summarizes the results of this experiment. As we can see,
for both \lmkl and \clmkl, using a scalable SVM solver greatly improves the
running time of the algorithm. In fact as we can see, the
methods using LibSVM fail to complete on certain inputs, whereas the methods
that use MWUMKL do not. We note that \textbf{MWUMKL} uses a parameter
$\epsilon$ which is the acceptable error in the duality gap of the SVM
optimization program. Higher $\epsilon$ values translate to more iterations, and
accuracy can often improve (up to a point) with lower $\epsilon$. Unless stated
otherwise, we use $\epsilon = 0.01$. Note that for this $\epsilon$, accuracy
does drop significantly in certain cases. 
 
\begin{figure}
  \includegraphics[width=\columnwidth]{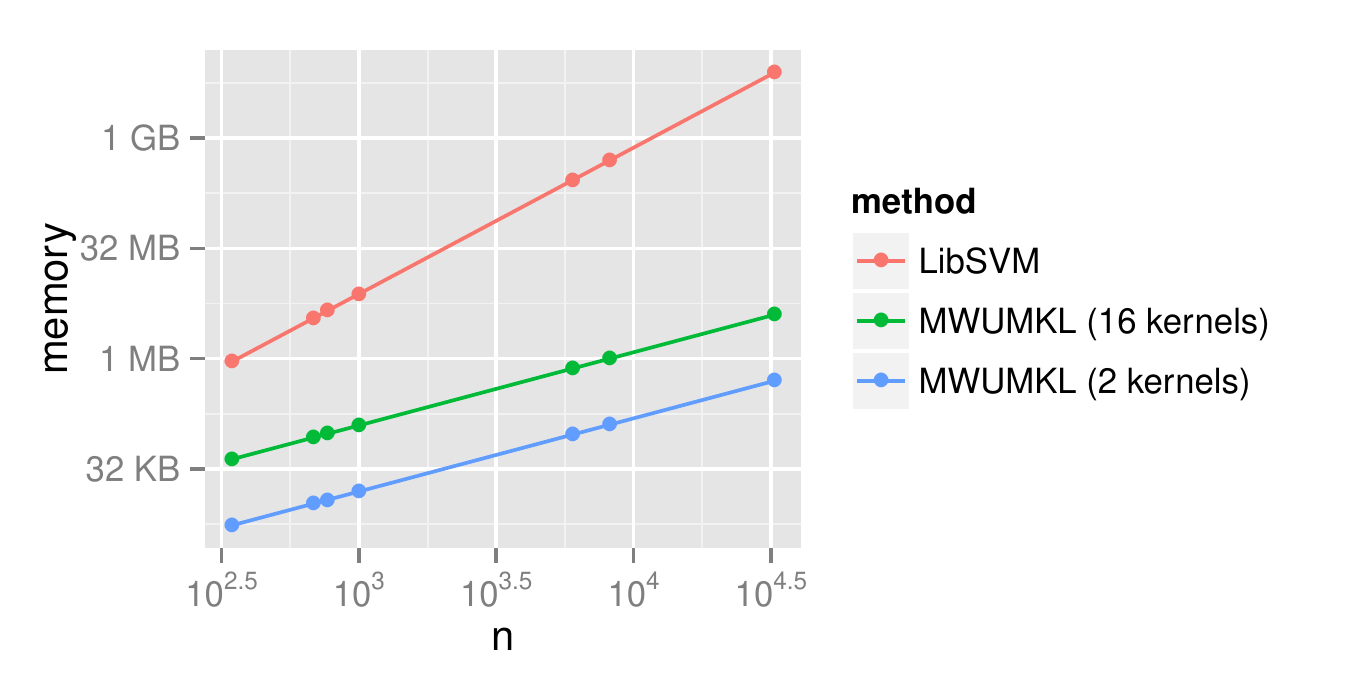}
  \caption{
    Minimum memory required (assuming double-precision floats) for LibSVM-based and MWUMKL-based methods. 
    LibSVM-based methods exclude those that use only LibSVM's standard kernels, such as \ldmkl, but include those that construct a new kernel, such as \lmkl, \clmkl, and \smkl.
    The values for $n$ are taken from the ``Examples'' column from Table~\ref{tab:t0-5}.
  }
  \label{fig:mem-scale}
\end{figure}

The case of \smkl is a little more interesting.  For smaller data sets the basic
method works quite well, and indeed outperforms any enhancement based on using
\textbf{MWUMKL}.  However, this comes at a price: the \smkl method requires a lot of
memory to solve the final kernel SVM with a kernel formed by combining the base
kernels.  For smaller data sets this effect does not materially affect
performance, but as we move to larger data sets like Adult, the method starts to
fail catastrophically. Figure~\ref{fig:mem-scale} illustrates the memory usage
incurred by the three localized methods when not using \textbf{MWUMKL} and when
using it. As we can see, the memory grows polynomially with the size of input. 

\paragraph{Stress-testing.}
Scaling \ldmkl to truly large datasets can present a challenge because we make use of kernelized support-vector regression.
There are several methods to address this problem which we will not enumerate here, but are targets for future versions of our algorithm.

\subsection{Support Points}
\label{sec:support-points}

We have argued earlier that localized multiple kernel learning methods have the
potential to generate classifiers with comparable accuracy but fewer support
points than global multiple kernel methods. This fact was first observed by \citet{DBLP:conf/icml/GonenA08}.
We now present detailed  empirical evidence establishing this claim. 
We compare the different localized kernel learning methods to \textsc{Uniform}
(a multiple kernel learning algorithm that merely takes an average of all the
kernels in its dictionary~\cite{corinna}), \gmkl~\cite{DBLP:conf/kdd/JainVV12}
(an iterative MKL solver that uses the spectral projected gradient), and
\textbf{MWUMKL}, run in its original form as a multiple kernel learning
algorithm. Results are presented in Table~\ref{tab:support}. 
While we did not annotate the results with accuracy numbers for ease of viewing, all methods have comparable accuracy (as Table~\ref{tab:t0} also indicates).

\begin{table*}[t]
  \footnotesize
  \centering
  \tabcolsep=0.15cm
  \begin{tabularx}{\textwidth}{lc c c cc c c} 
    \toprule
                   & \multicolumn{4}{c}{Localized Methods}                                    & \multicolumn{3}{c}{Global methods} \\
                   & \smkl            & \ldmkl         & \lmkl          &  \clmkl               & MWUMKL         & \textsc{Uniform} & GMKL \\
                   \cmidrule(r){2-5} \cmidrule(r){6-8}
    Breast         & 11.4\% (1\%)   & 12.9\% (1.1\%)   & 38\% (3.5\%) & \textbf{10.8\%} (1.1\%) & 21\% (1.4\%)   & 70.2\% (1.9\%) & 15.1\% (1\%) \\
    Diabetes       & \textbf{55.2\%} (1.3\%) & 56.4\% (1.3\%) & 58\% (1.7\%)   & 73.9\%  (10\%) & 79.1\% (1.7\%) & 70\% (2\%)     & 61.9\% (1.2\%) \\
    German         & 52.2\% (3.3\%) & \textbf{43.4\%} (2.4\%) & 89.2\% (2.8\%) & 99.8\% (0.3\%) & 81.7\% (1.1\%) & 60.8\% (1.6\%) & 68.4\% (1.4\%) \\
    Liver          & 82.2\% (1.7\%) & 70.2\% (7.3\%) & \textbf{63.1\%} (2.3\%) & 88.1\% (2.7\%) & 92.2\% (1.6\%) & 89.6\% (2.6\%) & 84.2\% (1.9\%) \\ 
    Mushrooms      & 4.3\% (0.2\%)  & 8.1\% (0.8\%)  & \textbf{1.9\%} (0.1\%)  & 4.0\% (0.3\%)  & 22.6\% (0.1\%) & 96.4\% (0.8\%) & 15.2\% (0.2\%) \\ 
    Gisette        & \textbf{20.8\%} (0.3\%) & 31.9\% (0.2\%) &  32.3\% (0.8\%)& 26.3\% (0.5\%) & 36.9\% (0.0\%) & 99.4\% (0.3\%) & 46.2\% (0.3\%) \\
    Adult          & 35.6\% (0.2\%) & 37.4\% (0.2\%) &   -                     & 35.4\% (0.2\%) & 40.4\% (0.0\%) & 48.2\% (0.2\%) & 41.7\% (0.1\%) \\ \bottomrule
  \end{tabularx}
  \caption{Numbers of support points computed as a percentage of the total number of points. 
    Numbers in parentheses are standard deviations over 100 iterations.  
    Values which are significantly superior to that of other methods are typeset in bold.}
  \label{tab:support}
\end{table*}

We observe that in all cases, the classifier using the fewest support points is always one of the localized methods, and the differences are always significant. 
However, it is not the case that a single local method always performs best. 
In general, \ldmkl (and \smkl)  appear to perform slightly better, but this is not consistent. 
Nevertheless, the results provide a clear justification for the argument that local kernel learning indeed finds sparser solutions.

\section{Related Work}

The general area of kernel learning was initiated by \citet{Lanckriet04MKLSDP} who proposed to simultaneously train an SVM as well as learn a convex combination of kernel functions. 
The key contribution was to frame the learning problem as an optimization over positive semidefinite kernel matrices which in turn reduces to a QCQP.
Soon after, \citet{DBLP:conf/icml/BachLJ04} proposed a block-norm regularization method based on \emph{second order cone programming} (SOCP). 

For efficiency, researchers started using optimization methods that alternate between updating the classifier parameters and the kernel weights. 
Many authors then explored the MKL landscape, including \citet{Sonnenburg06MKLSILP,Rakotomamonjy07MKL,DBLP:conf/nips/XuJKL08,DBLP:conf/icml/XuJYKL10}.
However, as pointed out in~\cite{corinna}, most of these methods do not compare favorably (both in accuracy as well as speed) even with the simple \emph{uniform} heuristic. More recently, \citet{DBLP:conf/aistats/MoellerRVS14} developed a multiplicative-weight-update based approach that has a much smaller memory footprint and scales far more effectively. 
Other \emph{global} kernel learning methods include  \cite{Micchelli:2005:LKF:1046920.1088710,DBLP:journals/jmlr/OngSW05,DBLP:conf/icml/VarmaB09,DBLP:conf/nips/CortesMR09,DBLP:conf/icml/OrabonaL11} and notably methods using the  $\ell_p$-norm~\cite{DBLP:journals/jmlr/KloftBSZ11,NIPS2009_0879,DBLP:conf/nips/VishwanathansAV10}.

\bibliographystyle{plainnat}
\bibliography{biblio2}

\end{document}